\documentclass{article} 
\usepackage{times,hyperref}
\usepackage{amssymb, amsbsy, amsmath}
\usepackage{xspace, multirow, url, graphicx}
\usepackage[numbers]{natbib}

\pdfoutput=1

\title{Towards Open-Text Semantic Parsing via Multi-Task Learning of Structured Embeddings}

 \author{
Antoine Bordes$^{(1,2)}$, Xavier Glorot$^{(2)}$, Jason Weston$^{(3)}$, Yoshua Bengio$^{(2)}$\\
{\small $(1)$ Heudiasyc, Universit\'e de Technologie de Compi\`egne, Compi\`egne, France}\hfill\\
{\small $(2)$ Dept. IRO, Universit\'e de Montr\'eal, Montr\'eal, QC, Canada}  \\
{\small $(3)$ Google, 111 8th Avenue, New York, NY, USA}\hfill
}
\date{}

\def\s{{\cal E}}
\def\lhs{$lhs$\xspace}
\def\rel{$rel$\xspace}
\def\rhs{$rhs$\xspace}

\def\lhslem{$lhs^{lem}$\xspace}
\def\rellem{$rel^{lem}$\xspace}
\def\rhslem{$rhs^{lem}$\xspace}

\def\lhssyn{$lhs^{syn}$\xspace}
\def\relsyn{$rel^{syn}$\xspace}
\def\rhssyn{$rhs^{syn}$\xspace}

\def\wordnet{WordNet\xspace}

\begin{document}

\maketitle

\begin{abstract}
  Open-text (or open-domain) semantic parsers are designed to
  interpret any statement in natural language by inferring a
  corresponding {\it meaning representation} (MR).
  Unfortunately, large scale systems cannot be easily machine-learned
  due to lack of directly supervised data.
  We propose here a method that learns to assign
  MRs 
  to a wide range of text (using a dictionary of more than 70,000
  words, which are mapped to more than 40,000 entities) thanks to a
  training scheme that combines learning from \wordnet and ConceptNet
  with learning from raw text. 
  The model learns structured embeddings of words, entities and MRs via a
  multi-task training process operating on these diverse sources of data that
  integrates all the learnt knowledge into a single system.
  This work ends up combining methods for knowledge acquisition, semantic
  parsing, and word-sense disambiguation.
  Experiments on various tasks indicate that our approach is indeed
  successful and can form a basis for future more sophisticated
  systems.

\end{abstract}

\section{Introduction}
{{\let\thefootnote\relax\footnotetext{Corresponding author: Antoine Bordes -- {\tt antoine.bordes@utc.fr} .}}}
A key ambition of AI has always been to render computers  able to
read text and express its meaning in a formal representation in
order to bring about a major improvement in human-computer
interfacing, question answering or knowledge acquisition.
Semantic parsing~\cite{mooney04} precisely aims at building such
systems to interpret statements expressed in natural language.
The purpose of the semantic parser is to analyze the structure
of sentence meaning and, formally, this consists of mapping a natural
language sentence into a logical {\it meaning representation} (MR).
This task seems too daunting to carry out manually (because of the vast
quantity of knowledge engineering that would be required) so machine
learning seems an appealing avenue.  On the other hand, machine
learning models usually require many labeled examples,
which can also be costly to gather, especially when labeling properly requires
the expertise of a linguist.

Hence, research in semantic parsing can be roughly divided in two
tracks.  
The first one, which could be termed {\it in-domain}, aims at learning
to build highly evolved and comprehensive MRs
~\cite{ge-mooney09,zettlemoyer09,liang11dcs}. Since this requires
advanced training data, such approaches have to be applied to text
from a specific domain with restricted vocabulary (a few hundred
words).
Alternatively, a second line of research, which could be termed {\it
  open-domain}, works towards learning to associate a MR to any kind of
natural language sentence~\cite{shi-mihalcea:2004, Giuglea:2006,
  poon-domingos:2009:EMNLP}. In this case, the supervision is much
weaker because it is unrealistic and infeasible to label data for
large-scale, open-domain semantic parsing. As a result, models usually
infer simpler MRs; this is sometimes referred to as {\it shallow}
semantic parsing.

In this paper, we propose a novel method directed towards the
open-domain category with the aim of automatically inducing
meaning representations out of free text, by exploiting
existing resources such as \wordnet to bootstrap and
anchor the process.
%
%
For a given sentence, the proposed approach infers a MR in two stages: (1) a
semantic role labeling (SRL) step predicts the semantic structure, and
(2) a disambiguation step assigns a corresponding entity to each
relevant word, so as to minimize an energy given to the whole input.

This paper considers simple MR structures and relies on an existing
method to perform SRL because its focus is on step (2). Indeed, in order
to go open-domain, a large number of entities must be considered. 
For this reason, the set of entities considered is defined from 
\wordnet~\cite{wordnet}. This results in a dictionary of more than
70,000 words that can be mapped to more than 40,000 possible entities.
%
For each word, \wordnet provides a list of candidate senses so step
(2) reduces to detecting the correct one and can be seen as
a challenging all-words word-sense disambiguation (WSD) step.
%

The model used here builds upon the {\em structured embedding framework}, defined
in \cite{bordesAAAI11}, in which each entity of a knowledge base
(such as \wordnet) is encoded into a low dimensional embedding vector space
preserving the original data structure.
%
The training procedure is based on multi-task learning across
different knowledge sources including \wordnet, ConceptNet~\cite{conceptnet} and raw text. In this
way MRs induced from raw text and MRs for \wordnet entities are
embedded (and hence integrated) in the same space. This allows us to
learn to perform disambiguation on raw text with {\em little direct and
  much indirect supervision}.  The model can learn to use \wordnet and ConceptNet
knowledge (such as relations between entities) to help choose the
correct sense of a particular word, and then label the words from
the raw text with the \wordnet sense.
%
%
At the same time, MR prediction can also be seen as knowledge extraction.
In addition to extracting MRs from raw text, the model proposed here
has the potential to enrich \wordnet with the extracted MRs.
The proposed method is evaluated on different criteria to reflect its different
properties. Thus, presented results illustrate MR inference,
word-sense disambiguation, \wordnet encoding and enrichment.

The paper is organized as follows. Section~\ref{sec:framework}
describes our framework to perform semantic
parsing. Section~\ref{sec:model} introduces our model based on
structured embeddings and Section~\ref{sec:training} the multi-task
training process we used to learn it. Section~\ref{sec:rwork}
discusses some related work. Finally our experiments are presented in
Section~\ref{sec:exp}.

\section{Semantic Parsing Framework} \label{sec:framework}

\subsection{Definitions} 
\label{sec:frame-def}

The MRs considered in semantic parsing are simple logical expressions
of the form $REL(A_0,\dots,A_n)$.  $REL$ is the relation symbol, and
$A_0$, ..., $A_n$ are its arguments. Note that several forms can be
recursively constructed to form more complex structures.
Because this work is oriented towards raw text, a wide
range of possible relation types and arguments must be considered.

Hence, \wordnet~\cite{wordnet} is used for defining the arguments and
some relation types as proposed in~\cite{shi-mihalcea:2004}.
\wordnet encompasses comprehensive knowledge within its graph
structure, whose nodes (termed {\it synsets}) correspond to senses, and
edges (which can have different types) define relations between those senses.
Each synset is associated with a set of words sharing that sense.
They are usually identified by 8-digit codes, however, for clarity
reasons, we indicate a synset by the concatenation of one of its
words, its part-of-speech tag and a number indicating which sense it
refers to (in the case of polysemous words).  For example, {\it
  \_score\_NN\_1} refers to the synset representing the first sense 
of the word ``score'' and also contains the words ``mark'' and
``grade'', whereas {\it \_score\_NN\_2} refers to the second meaning
(i.e. a written form of a musical composition).

We denote instances of relations from \wordnet using triplets (\lhs,
\rel, \rhs), where \lhs depicts the left-hand side of the relation,
\rel its type and \rhs its right-hand side.  Examples are ({\it
  \_score\_NN\_1} , {\it \_hypernym}, {\it \_evaluation\_NN\_1}) or
({\it \_score\_NN\_2} , {\it \_has\_part}, {\it
  \_musical\_notation\_NN\_1}).
In this work we filter out the synsets appearing in less that 15 triplets,
as well as relation types appearing in less than 5000 triplets. We obtain a
graph with the following statistics: 41,024 synsets and 18 relations
types; a total of 70,116 different words belong to these synsets.
%


\subsection{Methodology}~\label{sec:fram-prot}

\vspace*{-1cm}
\paragraph{MR structure inference (and preprocessing)}

The first stage consists in preprocessing the text and inferring the
structure of the MR.
Using the SENNA software\footnote{Freely available from
  \url{ml.nec-labs.com/senna/}.}~\cite{collobert11:arxiv}, we
performed part-of-speech (POS) tagging, chunking,
lemmatization\footnote{lemmatization is not carried out with SENNA but with the NLTK toolkit, \url{nltk.org}.} and semantic
role labeling (SRL).
The SRL step consists in labeling, for each proposition, each semantic
argument associated with a verb with its grammatical role. Each
argument is specified by a tuple of lemmas. 
It is crucial because it will be used to infer the {\em structure} of the
MR.
In this restricted setting, the structure of the MR follows that of the sentence.

We only consider sentences that match the following template: ({\em
  subject},\;{\em verb},\;{\em direct object}). Here, each of the three
elements of the template is associated with a tuple of lemmatized words
or synsets (when the words are disambiguated).
SRL is used to structure the sentence into the (\lhs = subject,
\rel = verb, \rhs = object) template, note that 
the order is not necessarily subject / verb / direct object in the raw
text. The semantic match energy
function is used to predict appropriate synsets or answer questions
by choosing those corresponding to low-energy synset configurations.

%

Clearly, the subject-verb-object structure causes the resulting MRs
to have a straightforward structure (with a single relation), but this
pattern is the most common and a good choice to test our ideas at
scale. Learning to infer more elaborate grammatical patterns is left
as future work.
In this work we chose to focus on handling the large scale of the set of entities.

As  an illustration, to parse the sentence: ``A musical score
accompanies a television program or a film.'', the SRL step will
produce as output the following triplet ({\it \_musical\_JJ}
{\it\_score\_NN}, {\it\_accompany\_VB}, {\it\_television\_program\_NN}
{\it\_film\_NN}).
In the following, we call the concatenation of a lemmatized word and
POS tag (such as NN, VB, etc.) a {\it lemma}. Note the absence of an
integer suffix, which distinguishes a lemma from a synset: a
lemma is allowed to be semantically ambiguous.
To summarize, this step starts from a sentence and either rejects it
or outputs a triplet of lemma tuples, one for the subject, one
for the relation or verb, and one for the direct object.

\paragraph{Detection of MR entities}

This second step aims at identifying each semantic entity expressed in
a sentence.
Given a relation triplet
(\lhslem, \rellem, \rhslem) where each element of the triplet is
associated with a tuple of lemmas, a corresponding triplet
(\lhssyn, \relsyn, \rhssyn) is produced, where the lemmas are replaced by synsets.
Depending on the lemmas, this can be either straightforward (some
lemmas such as {\it\_television\_program\_NN} or
{\it\_world\_war\_ii\_NN} correspond to a single synset) or very
challenging ({\it \_run\_VB} can be mapped to 41 different synsets and
{\it \_run\_NN} to 16).
Hence, in the proposed semantic parsing framework, MRs correspond to triplets of
synsets (\lhssyn, \relsyn, \rhssyn).
For the example from the previous section, the associated MR is (({\it
  \_musical\_JJ\_1},{\it\_score\_NN\_2}), {\it\_accompany\_VB\_1},
({\it\_television\_program\_NN\_1}, {\it\_film\_NN\_1})).

This step can be seen as particular form of all-words word-sense
disambiguation.
This is achieved by an approximate search for a set of synsets that are
compatible with the observed lemmas and that also minimize a semantic
matching energy function, using the model described in the next section.

Since the model is structured around relation triplets, MRs and
\wordnet relations are cast into the same scheme.  For example,
the \wordnet relation
 ({\it
  \_score\_NN\_2} , {\it \_has\_part}, {\it
  \_musical\_notation\_NN\_1}) fits the same pattern as our MRs, with the
relation type {\it \_has\_part} playing the role of
the verb.


\section{Structured Embeddings} \label{sec:model}


Inspired by the framework introduced by Bordes {\it et al}. \cite{bordesAAAI11} as well as by
recent work of L. Bottou \cite{tr-bottou-2011}, the main
idea behind our structural embedding model is the following.

\begin{itemize}
\item 
Named symbolic entities (including \wordnet synsets and relation types and lemmas) 
  are associated with a $d$-dimensional vector space,
  termed the ``embedding space'',
following previous work in neural language models~(see \cite{Bengio-scholarpedia-2007}
  for a review).
  The $i^{th}$ entity is assigned a
  vector $E_i \in {\mathbb R}^d$. 
  Note that if a lemma is unambiguous because it maps to a single synset, its
  embedding and the embedding of this synset are shared.
\item 
The semantic energy function value associated with a particular triplet 
(\lhs, \rel, \rhs) 
is computed by a parametrized function $\s$ that starts by mapping
all of the symbols to their embeddings. Note that in our case $\s$
must be able to handle variable-size arguments, since for example there
could be multiple lemmas in the subject part of the sentence.
\item
The energy function $\s$ is optimized to be lower for
training examples than for other possible configurations of symbols.
Hence the semantic energy function can distinguish plausible
combinations of entities from implausible ones, to choose the most
likely sense for a lemma, or to answer questions, e.g. corresponding to
a tuple $(lhs,rel,?)$ with a missing $rhs$ entry ``$?$''.
\end{itemize}

\subsection{Training Objective}

Let us now more formally define the training criterion for
the semantic match energy function. 
Let ${\cal C}$ denote the dictionary which includes all
entities (lemmas {\bf and} synsets) and relation types of interest, and let ${\cal C}^*$ denote
the set of tuples (or sequences) whose elements are taken in ${\cal C}$.
Let ${\cal R} \subset {\cal C}$ be the subset of entities which are
relation types (${\cal R}^*$ is defined similarly as ${\cal C}^*$).
We are given a training
set ${\cal D}$ containing $m$ triplets of the form $x=(x_{lhs},x_{rel},x_{rhs})$,
where $x_{lhs} \in {\cal C}^*$, $x_{rel} \in {\cal R}^*$, and $x_{rhs} \in {\cal C}^*$.
We define the energy as $\s(x)=\s(x_{lhs},x_{rel},x_{rhs})$. Ideally, we would like
to perform maximum likelihood over $P(x)\propto e^{-\s(x)}$ but this is intractable.
The approach we follow here has already been used successfully in ranking settings~\cite{unified_nlp,wsabie,usunier}
and corresponds to performing two approximations. First, like in pseudo-likelihood
we only consider one input given the others. Second, instead of sampling a negative
example from the model posterior, we use a ranking criterion (that is based on
uniformly sampling a negative example).

If one of the elements of a given triplet
were missing, then we would like the model to be able to predict the
correct entity.  For example, this would allow us to answer questions
like ``what is part of a car?'' or ``what does a score accompany?''.
The objective is to learn a real-valued semantic
energy function $\s$ 
such that it can successfully rank 
the training samples below all
other possible triplets:
\begin{align}
\label{con1}
\s(x) < \s(i,x_{rel},x_{rhs}) ~~\forall i \in {\cal C}^* : (i,x_{rel},x_{rhs}) \notin {\cal D}\\
\label{con2}
\s(x) < \s(x_{lhs},k,x_{rhs}) ~~\forall k \in {\cal R}^* : (x_{lhs},k,x_{rhs}) \notin {\cal D}\\
\label{con3}
\s(x) < \s(x_{lhs},x_{rel},j) ~~\forall j \in {\cal C}^* : (x_{lhs},x_{rel},j) \notin {\cal D}
\end{align}
In practice the following stochastic criterion is minimized:
\begin{equation}\label{crit}
\vspace*{-0.5mm}
   \sum_{x \in {\cal D}} \sum_{\tilde{x} \sim Q(\tilde{x}|x)} \max\left(\s(x)-\s(\tilde{x})+1,0\right)
\vspace*{-0.5mm}
\end{equation}
where $Q(\tilde{x}|x)$ is a corruption process
that transforms a training example $x$ into a corrupted {\em negative example}. 
In the experiments
$Q$ only changes one of the three members of the triplet, by changing only one of the
lemmas, synsets or relation type in it, by sampling it uniformly from $\cal C$
(not actually checking if the negative example is in ${\cal D}$).

\subsection{Parametrization of the Semantic Matching Energy Function}

\begin{figure}
  \begin{center}
    \includegraphics[width=0.5\linewidth]{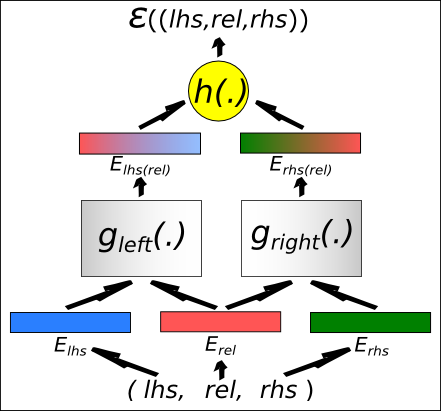}
    \caption{\label{fig:sem-nn} {\bf Semantic matching energy
        function.} {\small A triple of tuples $(lhs,rel,rhs)$ is first
        mapped to its embeddings $E_{lhs}$, $E_{lhs}$ and $E_{lhs}$
        (using an aggregating function for tuples involving more than
        one symbol). Then $E_{lhs}$ and $E_{rel}$ are combined using
        $g_{left}(.)$ to output $E_{lhs(rel)}$ (similarly
        $E_{rhs(rel)}=g_{right}(E_{rhs}, E_{rel})$). Finally the energy
        $\s((lhs,rel,rhs))$ is obtained by merging $E_{lhs(rel)}$ and $E_{rhs(rel)}$
        with the $h(.)$ function.}}
  \end{center}
\end{figure}

Many parametrizations are possible for the semantic matching energy function
but we have explored only a few. 
Let the input
triplet be $x$ $=$ $((lhs_1,lhs_2,\ldots),$ $(rel_1,rel_2,\ldots),$ $(rhs_1,rhs_2,\ldots))$.
In all of the experiments, the energy function
is structured as follows, based on the intuition that the relation type
should first be used to extract relevant components from each argument's embedding,
and put them in a space where they can then be compared (see Figure~\ref{fig:sem-nn} for an illustration).

\begin{itemize}
\item[(1)] Each symbol $i$ in the input tuples 
 is mapped to its embedding $E_i \in {\mathbb R}^d$.
\item[(2)] The embeddings associated with all the symbols within the same tuple
      are aggregated by a pooling function $\pi$ (we only used the mean in the
      experiments but other plausible candidates include the sum, the max,
      and combinations of several such elementwise statistics):
      \begin{align*}
        E_{lhs} = \pi(E_{lhs_1},E_{lhs_2},\ldots),\\
        E_{rel} = \pi(E_{rel_1},E_{rel_2},\ldots), \\ 
        E_{rhs} = \pi(E_{rhs_1},E_{rhs_2},\ldots),
      \end{align*}
where $lhs_i$ denotes the $i$-th individual element of the left-hand side tuple, etc.
\item[(3)] The embeddings $E_{lhs}$ and $E_{rel}$ respectively
 associated with the $lhs$ and $rel$ arguments
 are used to construct a new relation-dependent embedding $E_{lhs(rel)}$ for the $lhs$
 in the context of the relation type represented by $E_{rel}$,
 and similarly for the $rhs$:
       $E_{lhs(rel)} = g_{left}(E_{lhs},E_{rel})$ and
       $E_{rhs(rel)} = g_{right}(E_{rhs},E_{rel})$,
where $g_{left}$ and $g_{right}$ are parametrized functions whose parameters
are tuned during training. See more details in the experiments section.
\item[(4)] The energy is computed from the transformed embeddings of the left-hand side and right-hand side:
    $\s(x) = h(E_{lhs(rel)}, E_{rhs(rel)})$,
where $h$ is a parametrized function whose parameters
are tuned during training. More details are given in the experiments section.
\end{itemize}

\subsection{Disambiguation Process}~\label{sec:wsd}

Our semantic matching energy function is used for raw text semantic
to perform stage 2 of the protocol described in
Section~\ref{sec:fram-prot}, that is to carry out the word-sense
disambiguation step.

We label a triplet of lemmas
$((lhs^{lem}_1,lhs^{lem}_2,\ldots),(rel^{lem}_1,\ldots),(rhs^{lem}_1,\ldots))$
with synsets in a greedy fashion, one lemma at a time.
For labeling $lhs^{lem}_2$ for instance, we fix all the remaining elements of the triplet to their
lemma and select the synset leading to the lowest energy:
\begin{equation}
\vspace*{-0.5mm}
  lhs^{syn}_2=\text{argmin}_{S\in{\cal C}({syn|lem})}\s((lhs^{lem}_1,S,\ldots),(rel^{lem}_1,\ldots),(rhs^{lem}_1,\ldots))
\vspace*{-0.5mm}
\end{equation}
with ${\cal C}({syn|len})$ the set of allowed synsets to which $lhs^{lem}_2$ can
be mapped. We repeat that for all
lemmas. We always use lemmas as context, and never the already
assigned synsets.
%
%
This process is interesting because it is efficient as it only
requires to compute a low number of energies, equal to the number of
senses for a lemma.
However, it requires to have good representations (i.e. good embedding
vectors) for synsets and lemmas. That is the reason why the
multi-task training presented in next section, takes good care of learning both properly.

\section{Multi-Task Training} \label{sec:training} 

\subsection{Multiple Data Resources}

In order to encode as much common-sense knowledge as possible in the
model, the following heterogeneous data sources are combined.
\paragraph{\wordnet v3.0 (WN).}
Described in Section~\ref{sec:frame-def}, this is the main resource,
defining the dictionary of entities. The 18 relation types and 40,989
synsets retained are composed to form a total of 221,017 triplets.  We randomly extracted from them a
validation and a test set with 5,000 triplets each.

\wordnet contains only relations between synsets. However, the
disambiguation process needs embeddings for synsets and for lemmas.
Following~\cite{Havasi2010}, we created two other versions of this
dataset to leverage WN in order to also learn lemma embeddings: ``Ambiguated''
WN and ``Bridge'' WN. In ``Ambiguated'' WN both synset entities of each
triplet are replaced by one of their corresponding lemmas. ``Bridge'' WN
is designed to teach the model about the connection between synset and lemma
embeddings, thus in its relations the \lhs or \rhs synset is
replaced by a corresponding lemma.
Sampling training examples from WN involves actually sampling from one of its
three versions, resulting in a triplet involving synsets, lemmas or
both.

\paragraph{ConceptNet v2.1 (CN).} CN~\cite{conceptnet}  is a
common-sense knowledge base in which lemmas or groups of lemmas are
linked together with rich semantic relations as, for example, ({{\it
    \_kitchen\_table\_NN, \_used\_for, \_eat\_VB \_breakfast\_NN}).
  It is based on {\it lemmas} and not synsets, and it does not make
  distinctions between different senses of a word. Only triplets containing 
  lemmas from the WN dictionary are kept, to finally obtain a total of 11,332
  training lemma triplets.

\paragraph{Wikipedia (Wk).} 
This resource is simply raw text meant to provide knowledge to
the model in an unsupervised fashion.
In this work 50,000 Wikipedia articles were considered,
although many more could be used.
Using the protocol of the first paragraph of Section~\ref{sec:fram-prot},
we created a total of 1,484,966 triplets of lemmas.
Imperfect training triplets (containing a mix of lemmas and synsets) are produced by
performing the disambiguation step of Section~\ref{sec:wsd} on one of the
lemmas. This is equivalent to MAP (Maximum A Posteriori) training, i.e.,
we replace an unobserved latent variable by its mode according to a posterior
distribution (i.e. to the minimum of the energy function, given the observed variables).
We have used the 50,000 articles to generate more than 3M examples.

\paragraph{EXtended \wordnet (XWN) and Unambiguous Wikipedia (Wku).} 
XWN\cite{xwn} is built from \wordnet~{\it glosses}, syntactically
parsed and with content words semantically linked to WN synsets.
Using the protocol of Section~\ref{sec:fram-prot}, we processed these
sentences and collected 47,957 lemma triplets for which the synset MRs
were known.  We removed 5,000 of these examples to use them as an
evaluation set for the MR entity detection/word-sense disambiguation
task.
With the remaining 42,957 examples, we created unambiguous training
triplets to help the performance of the disambiguation algorithm described in
Section~\ref{sec:wsd}: for each lemma in each triplet, a new triplet is
created by replacing the lemma by its true corresponding synset and by keeping
the other members of the triplet in lemma form (to serve as examples of
lemma-based context).  This led to
a total of 786,105 training triplets, from which we removed 10,000 examples to
build a validation set.

We added to this training set some triplets extracted from the
Wikipedia corpus which were modified with the following trick: if one
of its lemmas corresponds unambiguously to a synset, and if this synset
maps to other ambiguous lemmas, we create a new triplet by replacing
the unambiguous lemma by an ambiguous one. Hence, we know the true
synset in that ambiguous context. This allowed to create 981,841
additional triplets with supervision, and we named this data set
Unambiguous Wikipedia.

\subsection{Training Procedure}

To train the parameters of the energy function $\s$ we loop over all of
the training data resources and use stochastic gradient descent (SGD)
\cite{robbins_monro:1951}. That is, we iterate the following steps:
\iftrue
\begin{enumerate}
\vspace*{-0.02in}
\item Select a positive training triplet $x_i$ at random (composed of synsets, of lemmas or both)
from one of the above sources of examples.
\vspace*{-0.02in}
\item Select at random resp. constraint~\eqref{con1},~\eqref{con2}
    or~\eqref{con3}.
\vspace*{-0.02in}
\item Create a negative triplet $\tilde{x}$ by sampling an entity
    from $\cal C$ to replace resp. $lhs_i$, $rel_i$ or
    $rhs_i$.   \label{step:sample}
\vspace*{-0.02in}
\item If $\s(x_i) > \s(\tilde{x})  - 1$, make a stochastic gradient step to minimize the criterion~\eqref{crit}. \label{step:margin}
\vspace*{-0.1in}
\item Enforce the constraint that each embedding vector is normalized, $||E_i|| = 1$, $\forall i$. \label{step:normalization}
\end{enumerate}

\else
\begin{enumerate}
\vspace*{-0.02in}
\item Select a positive training triplet $x_i$ at random (composed of synsets, of lemmas or both)
from one of the above sources of examples.
\vspace*{-0.02in}
\item Select at random resp. constraint~\eqref{con1},~\eqref{con2}
    or~\eqref{con3} and set N=0.
\vspace*{-0.02in}
\item Repeat:
{\small
\vspace*{-0.02in}
  \begin{enumerate}
  \item Create a negative triplet $\tilde{x}$ by sampling an entity
    from $\cal C$ to replace resp. $lhs_i$, $rel_i$ or
    $rhs_i$.    
\vspace*{-0.02in}
  \item N=N+1
  \end{enumerate}
}
\vspace*{-0.02in}
until $\s(x_i) > \s(\tilde{x})  - 1$ or $N \geq \gamma$.
\vspace*{-0.02in}
\item Make a stochastic gradient step to minimize the criterion~\eqref{crit}.
\vspace*{-0.02in}
\item Enforce the constraint that each embedding vector is normalized, $||E_i|| = 1$, $\forall i$.
\end{enumerate}
The loop in step 3 allows to seamlessly optimize a WARP loss
(Weighted Approximate-Rank Pairwise loss) which has shown to enhance
training of embedding-based models~\cite{wsabie}. The parameter
$\gamma$ controls the number of sampling steps (100 in our experiments).
\fi
The constant $1$ in step~\ref{step:margin} is the {\bf margin} as is commonly
used in many margin-based models such as SVMs \cite{boser1992tao}.
The gradient step requires a learning rate of $\lambda$.  The
normalization in step~\ref{step:normalization} helps remove scaling freedoms from the
model.

The above algorithm was used for all the data sources
except XWN and Wku. In that case, positive triplets are composed of lemmas
(as context) and of a disambiguated lemma replaced by its synset. Unlike
for Wikipedia, this is labeled data, so we are certain that this
synset is the true sense.
Hence, to increase training efficiency and yield a more discriminant 
disambiguation, in step~\ref{step:sample} with probability $\frac{1}{2}$ we
either sample randomly from $\cal C$ or we sample randomly from the
set of remaining candidate synsets corresponding to this disambiguated
lemma (i.e. the set of its other meanings).

The matrix $E$ which contains the representations of the entities is thus
learnt via a complex {\em multi-task learning} procedure because a single
embedding matrix is used for all relations and all data sources
(each really corresponding to a different distribution of symbol tuples,
i.e., a different task).
As a result, the embedding of an entity contains factorized information coming from
all the relations in which the entity is involved as \lhs, \rhs or
even \rel (for verbs).
For each entity, the model is forced to learn how it interacts
with other entities in many different ways.

\section{Related Work} \label{sec:rwork}

Our approach is original by the way that it connects many tasks and
many training resources within the same framework. However, it is
highly related with many previous works.
%
Shi and Mihalcea~\cite{shi-mihalcea:2004} proposed a rule-based system
for open-text semantic parsing using \wordnet and
FrameNet~\cite{framenet} while Giuglea and Moschitti
\cite{Giuglea:2006} proposed a model to connect \wordnet, VerbNet and
PropBank~\cite{propbank}
 for semantic parsing using tree kernels.
Poon and
Domingos~\cite{poon-domingos:2009:EMNLP,poon-domingos:2010:ACL}
recently introduced a method based on Markov-Logic Networks for
unsupervised semantic parsing that can be also used for information
acquisition. However, instead of connecting MRs to an existing
ontology the proposed method does, it constructs a new one and does not leverage
pre-existing knowledge.
Automatic information extraction is the topic of many models and
demos~\cite{Snow:2006, yates-EtAl:2007, wu-weld:2010:ACL, Suchanek:2008} 
 but none of them relies on a joint embedding model.
In that trend, some approaches have been directly targeting to enrich
existing resources, as we do here with \wordnet,~\cite{Agirre00,
  cuadrosrigau08a, Cimiano:2006} 
 but these never use learning.
Finally, several previous works have targeted to improve WSD
by using extra-knowledge by either automatically acquiring
examples~\cite{Martinez:2008} or by connecting different knowledge
bases~\cite{Havasi2010}.

Our model is related to earlier approaches~(e.g. \cite{bengio03,
  unified_nlp,Paccanaro:2001,Cambria:2009}) and is similar to but more convenient
than the approach of \citet{bordesAAAI11}, where the embeddings for
the left/right-hand side arguments $i$ and $j$ are
$d$-vectors and the embedding for relation $k$ is a pair of
$d \times d$ matrices.
The disadvantage of embedding each relation type into a pair of matrices is that it gives relation
types a different status (they cannot appear as left-hand side or
right-hand side) and many more parameters.  



\section{Experiments} \label{sec:exp}

\subsection{Experimental Setting}

Experiments were performed with three different types of
parametrizations (linear, bilinear and non-linear) for the $g$
functions. We selected the hyper-parameter values w.r.t.
the WN validation set. We only present in this section
results with the bilinear parametrization for $g$ and with a dot product for
the output $h$ function, $h(a,b)=a\cdot b$, because this combination achieved the best 
performance in validation. The bilinear function for the left side
(and similarly for the right side) is as follows:\\
{{\small
$$
g_{left}(E_{lhs},E_{rel})^l = \sum_{i,j,k} (W_{3 left}^{kl} * (W_{1 left}^{ik}*E_{lhs}^{i}+b_{1 left}^{k}) * (W_{2 left}^{jk}*E_{rel}^{j}+b_{2 left}^{k}) + b_{3 left}^{l})
$$
}}%
where $i$ denotes indices of the elements of the embedding $E_{lhs}$, $j$ of the relation embedding $E_{rel}$, $k$ of the latent 
representation, and $l$ of the output of the $g$ function (that will be fed into the dot product).
We hypothesize that the success of the bilinear parametrization comes from its natural ability to
encode AND relationships between the \lhs (or \rhs) and the \rel embeddings.

To assess the performance w.r.t. choices made with
that architecture, the multi-task training and the diverse data
sources, we evaluated models trained with several combinations of data sources.
{\bf WN} denotes models trained on \wordnet, ``Ambiguated'' \wordnet
and ``Bridge'' \wordnet, {\bf WN+CN+Wk} models also trained
on CN and Wk datasets, and {\bf All} models
are trained on all sources. 
 
\begin{table}[t]
  \caption{{\bf \wordnet encoding and Word Sense Disambiguation results.}
   MFS is just using the Most Frequent Sense. {\bf All+MFS} is our best system, combining
   all sources of information. {\bf Random} chooses uniformly among allowed synsets.
    {\small $(^*)$Results of {\bf StructEmbed}, copied from \cite{bordesAAAI11}, were obtained with a 
      different version of \wordnet and are presented here as indication only.}} 
    \begin{center}
      {\small
    \begin{tabular}{|l|c|c||c|c|}\hline
        {\bf Model} & {\bf \wordnet rank} & {\bf \wordnet top-10}  & {\bf F1 XWN} & {\bf F1 Senseval3} \\ \hline
        {\bf All+MFS}  & --          & --            & {\bf 72.35}\%  & {\bf 70.19}\% \\ \hline
        {\bf All}        & 139.3       & 34.71\%       & 67.52\%        & 51.44\% \\ \hline
        {\bf WN+CN+Wk}   & 95.9        & 46.02\%       & 34.80\%        & 34.13\% \\ \hline
        {\bf WN}         & {\bf 72.1}  & {\bf 58.87}\%       & 29.55\%        & 28.36\% \\ \hline \hline
        {\bf MFS}        & --          & --            & 67.17\%        & 67.79\% \\ \hline
        {\bf Gamble} \cite{Gamble} & --       & -- & --             & 66.41\% \\ \hline
        {\bf StructEmbed} \cite{bordesAAAI11} & \it 140.1$^*$       & \it 74.20\%$^*$ & --  & -- \\ \hline
        {\bf Random}     & 20512       & 0.024\%       & 26.71\%        & 29.55\% \\ \hline
    \end{tabular}
  }
    \end{center}
\label{tab:WSD-WN}
\end{table}

\subsection{\wordnet Encoding}\label{sec:wn-trip}

The WN encoding is measured with the mean predicted rank and 
the prediction at top 10 (top-10), calculated with 
the following procedure.
For each test \wordnet triplet, the left entity is removed and
replaced by each of the 41,024 synsets of the dictionary in
turn. Energies of those degraded triplets are computed by the model
and sorted by ascending order and the rank of the correct synset is
stored. That is done for both the left-hand and right-hand
arguments of the relation.  The mean predicted rank is the average of
those predicted ranks and top-10 is the proportion of ranks within 1
and 10.

The left side of Table~\ref{tab:WSD-WN} presents the comparative results, together with
previous performance of~\cite{bordesAAAI11} ({\bf StructEmbed}). We obtain better
predicted rank but lower top-10. However, the difference can be caused
by the fact that the dictionaries of synsets are slightly different
between ~\cite{bordesAAAI11} and the setting presented here 
(different preprocessing of WordNet).
Training with other data sources ({\bf All}) still allows to encode
well \wordnet knowledge, even if it is slightly worse than with \wordnet alone ({\bf WN}).

\subsection{Word Sense Disambiguation}\label{sec:exp-wsd}

Performance on WSD is assessed on two test sets: the XWN test set and
a subset of English All-words WSD task of
SensEval-3.\footnote{\url{www.senseval.org/senseval3}.}  For the latter, we
processed the original data using the protocol of
Section~\ref{sec:fram-prot} and obtained a total of 208 words to
disambiguate (out of $\approx 2000$ originally). The performance of
the most frequent sense ({\bf MFS}) based on \wordnet frequencies is also
evaluated. Finally, we also report the results of {\bf
  Gamble}~\cite{Gamble}, winner of Senseval-3, on our subset of its data.

F1 scores are presented in Table~\ref{tab:WSD-WN} (right). The
difference between {\bf All} and {\bf WN+CN +Wk} indicates that, even
without direct supervision, the model can disambiguate some words
({\bf WN+CN+Wk} is significantly above {\bf Random}),
that the information from XWN and Wku is crucial (+30\%) and yields
performance better than {\bf MFS} (a strong baseline in WSD) on the
XWN test set.

However, performance can be greatly improved by combining the {\bf All} sources
model and the MFS score.
To do so, we converted the frequency information into an energy by
taking minus the log frequency and used it as an extra
energy term. The total energy function is used for disambiguation.
This yields the results denoted by {\bf All+MFS} which achieves the
best performance of all the methods tried.

\begin{figure}
  \begin{center}
    \includegraphics[width=0.6\linewidth]{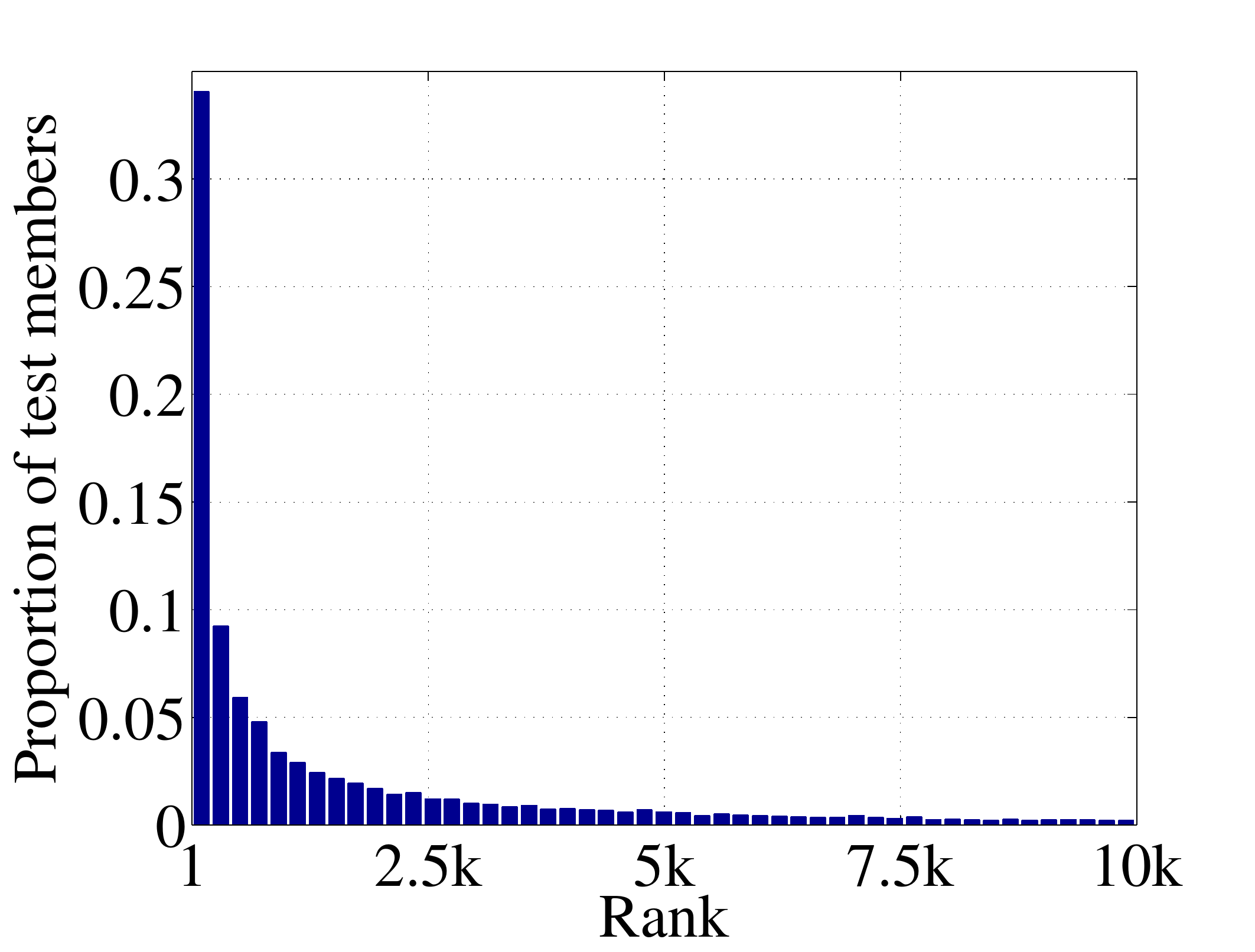}
    \caption{\label{fig:rank} {\bf MR ranking with respect
      to the model energy on the XWN test set.} {\small On a total of 41,024 synset entities,
      the median/mean rank is 640/3012 whereas with \wordnet::Similarity we obtained 13,200/13,800.}
    }
  \end{center}
\end{figure}

\subsection{Meaning Representations}

Assessing the quality of the obtained MRs is difficult as there is no
benchmark for open-text semantic parsing.
Yet, we intend to give an insight of how well the model
represents language information by measuring and ranking the energy 
function of MRs.
We measure the ranks of the predicted and the correct MRs obtained
from the XWN test set, like \wordnet
triplets were treated in Section~\ref{sec:wn-trip}: we replace each lemma of a
triplet by all the 41,024 synsets to rank (by energy) both correct and predicted synsets
(according to {\bf All+MFS}).
In both cases we obtain comparable median (mean) ranks: 640 (3012)/41,024 for 
the correct representation and 516 (2443)/41,024 for the prediction.
Figure~\ref{fig:rank} shows a histogram of the ranks of each
correct MR of the test set for {\bf All+MFS}.
These low ranks indicates that the model has learnt to give lower
energies to plausible MRs, i.e. has integrated higher level
information about how synsets and lemmas can form sentences in language.

Interestingly, the \wordnet::Similarity package~\cite{WN_similarity}
can also be used to rank MRs because it is designed to compute a
similarity between synsets using the \wordnet graph. We used the
package's  second order co-occurrence vector of synset definitions
to compute similarity, which gave the best results.
This leads to 13,500 (14,000)/41,024 median (mean) ranks for the
correct representations.
Even though, these ranks are above chance, they are far worse than
those of our model.
This is mainly caused by the fact that \wordnet::Similarity only knows
about relations between language entities through the \wordnet graph, for
which the number of relation type is very low ($\approx 20$).
For instance, it has no clue that a noun and a verb can (and how they
can) be related, contrary to our model that learns that through its
multi-task training on raw text.



\begin{table}
  \begin{center}
    {\small
      \begin{tabular}{|c|c|c|}
        \hline
        & Model (All)  & TextRunner \\
        \hline
        \lhs  & \_army\_NN\_1   & army \\
        \hline
        \rel  & \_attack\_VB\_1 & attacked \\
        \hline
        & \_troop\_NN\_4               & Israel       \\
        top    & \_armed\_service\_NN\_1      & the village  \\
        ranked & \_ship\_NN\_1                & another army \\
        \rhs   & \_territory\_NN\_1           &  the city     \\
        & \_military\_unit\_NN\_1      & the fort     \\
        \hline
        \hline
        & \_business\_firm\_NN\_1  & People           \\
        top          & \_person\_NN\_1          & Players          \\
        ranked       & \_family\_NN\_1          & one              \\
        \lhs         & \_payoff\_NN\_3          & Students         \\
        & \_card\_game\_NN\_1      & business         \\
        \hline
        \rel        & \_earn\_VB\_1 & earn  \\
        \hline
        \rhs        & \_money\_NN\_1 & money \\
        \hline
      \end{tabular}
    }
    \caption{{\bf Lists of entities} {\small reported by our system and by TextRunner.}\label{tab:lists}}
    
  \end{center}
\end{table}

\subsection{\wordnet Enrichment}

\wordnet and ConceptNet use a limited number of relation types.
Thanks to its multi-task training and its unified representation for MRs
and \wordnet/ConceptNet relations, our model is able to learn rich
relationships between synsets and lemmas and can even enrich
those knowledge bases since it sees every verb as a potential
relation type.
Therefore our model is able to learn richer relationships between
synsets and lemmas entities.
As illustration, predicted lists for relation types that do not exist in the
two knowledge bases are given in Table~\ref{tab:lists}. We also
compare with lists returned by TextRunner~\cite{yates-EtAl:2007} (an
information extraction tool having extracted information from 100M webpages, to be compared
with our 50k Wikipedia articles).
Lists from both systems truly reflect common-sense. However, contrary to our system,
TextRunner does not disambiguate different senses of a lemma,
and thus it cannot connect its knowledge to an existing resource
to enrich it.

\section{Conclusion} \label{sec:concl}

In this work we developed a large-scale system for semantic parsing from raw text
to disambiguated meaning representations.
The generalization ability of our method crucially centers upon 
 scoring triplets of  relations between ambiguous lemmas and unambiguous concepts (synsets) 
both using a single structured embedding energy function.
Multi-tasking the learning of such a function
 over several resources we effectively learn to build
disambiguated meaning representations from raw text with little direct supervision.        

The final system can potentially capture the deep semantics of sentences 
in the structured embedding energy function
by generalizing
the knowledge learnt across the multiple resources (e.g. common-sense knowledge from ConceptNet
and relations between concepts from WordNet) and linking it to raw text (from Wikipedia).
We obtained positive experimental results on several semantic tasks that appear to support this assertion,
 but future work should explore the capabilities of such systems further including other semantic tasks,
and utilizing more evolved grammars, e.g. by using FrameNet~\cite{framenet} (see e.g.~\cite{Coppola:2010}).


\subsection*{Acknowledgments}
The authors would like to acknowledge L\'eon Bottou, Nicolas Usunier and
Ronan Collobert for inspiring discussions.  This work was supported by
DARPA DL Program, CRSNG, MITACS, RQCHP and SHARCNET. All experimental
code has been implemented using the Theano
library~\citep{bergstra+al:2010-scipy}.

\bibliography{aigaion,semantic_parsing,aaai_paper}
\bibliographystyle{natbib}

\end{document}